# Embodied Artificial Intelligence through Distributed Adaptive Control: An Integrated Framework

Clément Moulin-Frier
SPECS Lab
Universitat Pompeu Fabra
Barcelona, Spain
Email: clement.moulinfrier@gmail.com

Jordi-Ysard Puigbò
SPECS Lab
Universitat Pompeu Fabra
Barcelona, Spain
Email: jordiysard.puigbo@upf.edu

Xerxes D. Arsiwalla
SPECS Lab
Universitat Pompeu Fabra
Barcelona, Spain
Email: x.d.arsiwalla@gmail.com

Martì Sanchez-Fibla
SPECS Lab
Universitat Pompeu Fabra
Barcelona, Spain
Email: santmarti@gmail.com

Paul FMJ Verschure
SPECS Lab
Universitat Pompeu Fabra
ICREA
Institute for Bioengineering of Catalonia (IBEC)
Barcelona Institute of Science and Technology (BIST)
Barcelona, Spain
Email: paul.verschure@upf.edu

*Abstract*—In this paper, we argue that the future of Artificial Intelligence research resides in two keywords: *integration* and *embodiment*. We support this claim by analyzing the recent advances in the field. Regarding integration, we note that the most impactful recent contributions have been made possible through the integration of recent Machine Learning methods (based in particular on Deep Learning and Recurrent Neural Networks) with more traditional ones (e.g. Monte-Carlo tree search, goal babbling exploration or addressable memory systems). Regarding embodiment, we note that the traditional benchmark tasks (e.g. visual classification or board games) are becoming obsolete as state-of-the-art learning algorithms approach or even surpass human performance in most of them, having recently encouraged the development of first-person 3D game platforms embedding realistic physics. Building on this analysis, we first propose an embodied cognitive architecture integrating heterogeneous subfields of Artificial Intelligence into a unified framework. We demonstrate the utility of our approach by showing how major contributions of the field can be expressed within the proposed framework. We then claim that benchmarking environments need to reproduce ecologically-valid conditions for bootstrapping the acquisition of increasingly complex cognitive skills through the concept of a cognitive arms race between embodied agents.

## I. INTRODUCTION

In recent years, research in Artificial Intelligence has been primarily dominated by impressive advances in Machine Learning, with a strong emphasis on the so-called Deep Learning framework. It has allowed considerable achievements such as human-level performance in visual classification [1] and description [2], in Atari video games [3] and even in the highly complex game of Go [4]. The Deep Learning approach is characterized by supposing very minimal prior on the task to be solved, compensating this lack of prior knowledge by feeding the learning algorithm with an extremely high amount of training data, while hiding the intermediary representations.

However, it is important noting that the most important contributions of Deep Learning for Artificial Intelligence often owe their success in part to their *integration* with other types of learning algorithms. For example, the AlphaGo program which defeated the world champions in the famously complex game of Go [4], is based on the integration of Deep Reinforcement Learning with a Monte-Carlo tree search algorithm. Without the tree search addition, AlphaGo still outperforms previous machine performances but is unable to beat high-level human players. Another example can be found in the original Deep Q-Learning algorithm (DQN, Mnih et al., 2015), achieving very poor performance in some Atari games where the reward is considerably sparse and delayed (e.g. Montezuma Revenge). Solving such tasks has required the integration of DQN with intrinsically motivated learning algorithms for novelty detection [5], or goal babbling [6].

A drastically different approach has also received considerable attention, arguing that deep learning systems are not able to solve key aspects of human cognition [7]. The approach states that human cognition relies on building causal models of the world through combinatorial processes to rapidly acquire knowledge and generalize it to new tasks and situations. This has led to important contributions through model-based Bayesian learning algorithms, which surpass deep learning approaches in visual classification tasks while displaying powerful generalization abilities in one-shot training [8]. This solution, however, comes at a cost: the underlying algorithm requires a priori knowledge about the primitives to learn from and about how to compose them to build increasingly abstract categories. An assumption of such models is that learning should be grounded in intuitive theories of physics and psychology, supporting and enriching acquired knowledge [7], as supported by infant behavioral data [9].

Considering the pre-existence of intuitive physics and psychology engines as an inductive bias for Machine Learning is far from being a trivial assumption. It immediately raises the question: where does such knowledge come from and how is it shaped through evolutionary, developmental and cultural processes? All the aforementioned approaches are lacking this fundamental component shaping intelligence in the biological world, namely *embodiment*. Playing Atari video games, complex board games or classifying visual images at a human level are considerable milestones of Artificial Intelligence research. Yet, in contrast, biological cognitive systems are intrinsically shaped by their physical nature. They are embodied within a dynamical environment and strongly coupled with other physical and cognitive systems through complex feedback loops operating at different scales: physical, sensorimotor, cognitive, social, cultural and evolutionary. Nevertheless, many recent Artificial Intelligence benchmarks have focused on solving video games or board games, adopting a third-person view and relying on a discrete set of actions with no or poor environmental dynamics. A few interesting software tools have however recently been released to provide more realistic benchmarking environments. This for example, is the case of Project Malmo [10] which provides an API to control characters in the *MineCraft* video game, an open-ended environment with complex physical and environmental dynamics; or *Deepmind Lab* [11], allowing the creation of rich 3D environments with similar features. Another example is *OpenAI Gym* [12], providing access to a variety of simulation environments for the benchmarking of learning algorithms, especially reinforcement learning based. Such complex environments are becoming necessary to validate the full potential of modern Artificial Intelligence research, in an era where human performance is being achieved on an increasing number of traditional benchmarks. There is also a renewed interest for multi-agent benchmarks in light of the recent advances in the field, solving social tasks such as the prisoner dilemma [13] and studying the emergence of cooperation and competition among agents [14].

The above examples emphasize two important challenges in modern Artificial Intelligence. Firstly, there is a need for a unified integrative framework providing a principled methodology for organizing the interactions of various subfields (e.g. planning and decision making, abstraction, classification, reinforcement learning, sensorimotor control or exploration). Secondly, Artificial Intelligence is arriving at a level of maturation where more realistic benchmarking environments are required, for two reasons: validating the full potential of the state-of-the-art artificial cognitive systems, as well as understanding the role of environmental complexity in the shaping of cognitive complexity.

In this paper, we first propose an embodied cognitive architecture structuring the main subfields of Artificial Intelligence research into an integrated framework. We demonstrate the utility of our approach by showing how major contributions of the field can be expressed within the proposed framework, providing a powerful tool for their conceptual description and comparison. Then we argue that the complexity of a cognitive agent strongly depends on the complexity of the environment it lives in. We propose the concept of a cognitive arms race, where an ecology of embodied cognitive agents interact in a dynamic environment reproducing ecologically-valid conditions and driving them to acquire increasingly complex cognitive abilities in a positive feedback loop.

## II. AN INTEGRATED COGNITIVE ARCHITECTURE FOR EMBODIED ARTIFICIAL INTELLIGENCE

Considering an integrative and embodied approach to Artificial Intelligence requires dealing with heterogeneous aspects of cognition, where low-level interaction with the environment interacts bidirectionally with high-level reasoning abilities. This reflects a historical challenge in formalizing how cognitive functions arise in an individual agent from the interaction of interconnected information processing modules structured in a cognitive architecture [15], [16]. On one hand, top-down approaches mostly rely on methods from Symbolic Artificial Intelligence (from the General Problem Solver [17] to Soar [18] or ACT-R [19] and their follow-up), where a complex representation of a task is recursively decomposed into simpler elements. On the other hand, bottom-up approaches instead emphasize lower-level sensory-motor control loops as a starting point of behavioral complexity, which can be further extended by combining multiple control loops together, as implemented in behavior-based robotics [20] (sometimes referred as *intelligence without representation* [21]). These two approaches thus reflect different aspects of cognition: high-level symbolic reasoning for the former and low-level embodied behaviors for the latter. However, both aspects are of equal importance when it comes to defining a unified theory of cognition. It is therefore a major challenge of cognitive science to unify both approaches into a single theory, where (a) reactive control allows an initial level of complexity in the interaction between an embodied agent and its environment and (b) this interaction provides the basis for learning higher-level representations and for sequencing them in a causal way for top-down goal-oriented control.

For this aim, we adopt the principles of the Distributed Adaptive Control (DAC) theory of the mind and brain [22], [23]. Besides its biological grounding, DAC is an adequate modeling framework for integrating heterogeneous concepts of Artificial Intelligence and Machine Learning into a coherent cognitive architecture, for two reasons: (a) it integrates the principles of both the aforementioned bottom-up and top-down approaches into a coherent information processing circuit; (b) it is agnostic to the actual implementation of each of its functional modules. Over the last fifteen years, DAC has been applied to a variety of complex and embodied benchmark tasks, for example foraging [22], [24] or social humanoid robot control [16], [25].

*A. The DAC-EAI cognitive architecture: Distributed Adaptive Control for Embodied Artificial Intelligence*

DAC posits that cognition is based on the interaction of interconnected control loops operating at different levels of abstraction (**Figure 1**). The functional modules constituting the architecture are usually described in biological or psychological terms (see e.g. [26]). Here we propose instead to describe them in purely computational term, with the aim of facilitating the description of existing Artificial Intelligence systems within this unified framework. We call this instantiation of the architecture *DAC-EAI*: Distributive Adaptive Control for Embodied Artificial Intelligence.

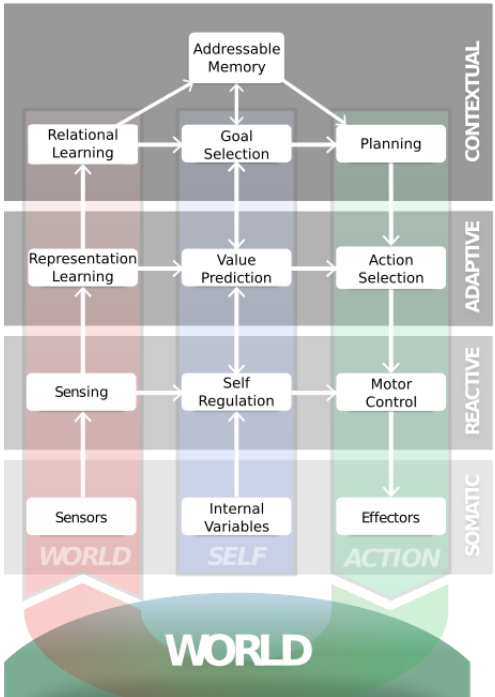

**Figure 1: The DAC-EAI architecture allows a coherent organization of heterogeneous subfields of Artificial Intelligence. DAC-EAI stands for Distributed Adaptive Control for Embodied Machine Learning.** It is composed of three layers operating in parallel and at different levels of abstraction. See text for detail, where each module name is referred with *italics*.

The first level, called the Somatic layer, corresponds to the embodiment of the agent within its environment. It includes the sensors and actuators, as well internal variables to be regulated (e.g. energy or safety levels). The self-regulation of these internal variables occurs in the Reactive layer and extends the aforementioned behavior-based approaches (e.g. the Subsumption architecture [20]) with drive reduction mechanisms through predefined sensorimotor control loops (i.e. reflexes). In **Figure 1**, this corresponds to the mapping from the *Sensing* to the *Motor Control* module through *Self Regulation*. The Reactive layer offers several advantages when analyzed from the embodied artificial intelligence perspective of this paper. First, reward is traditionally considered in Machine Learning as a scalar value associated with external states of the environment. DAC proposes instead that it should derive from the internal dynamics of multiple internal variables modulated by the body-environment real-time interaction, providing an embodied notion of reward in cognitive agents. Second, the Reactive layer generates a first level of behavioral complexity through the interaction of predefined sensorimotor control loops for self-regulation. This provides a notion of embodied inductive bias bootstrapping and structuring learning processes in the upper levels of the architecture. This is a departure from the model-based approaches mentioned in the introduction [7], where inductive biases are instead considered as intuitive core knowledge in the form of a pre-existent physics and psychology engine.

Behavior generated in the Reactive layer bootstraps learning processes for acquiring a state space of the agent-environment interaction in the Adaptive layer. The *Representation Learning* module receives input from *Sensing* to form increasingly abstract representations. For example, unsupervised learning methods such as deep autoencoders [27] could be a possible implementation of this module. The resulting abstract states of the world are mapped to their associated values through the *Value Prediction* module, informed by the internal states of the agent from *Self Regulation*. This allows the inference of action policies maximizing value through *Action Selection*, a typical reinforcement learning problem [28]. We note that Deep Q-Learning [3] provides an integrated solution to the three processes involved in the Adaptive layer, based on Deep Convolutional Networks for *Representation Learning*, Q-value estimation for *Value Prediction* and an ε-greedy policy for *Action Selection*. However, within our proposed framework, the self-regulation of multiple internal variables in the Reactive layer requires the agent to switch between different action policies (differentiating e.g. between situation of low energy vs. low safety). A possible way to achieve this using the Deep Q-Learning framework is to extend it to multi-task learning (see e.g. [29]). Since it is likely that similar abstract features are relevant to various tasks, a promising solution is to share the representation learning part of the network (the convolutional layers in [3]) across tasks, while multiplying the fully-connected layers in a task-specific way.

The state space acquired in the Adaptive layer then supports the acquisition of higher-level cognitive abilities such as goal selection, memory and planning in the Contextual layer. The abstract representations acquired in *Representation Learning* are linked together through *Relational Learning*. The availability of abstract representations in possibly multiple modalities provides the substrate for causal and compositional linking. Several state-of-the-art methods are of interest for learning such relations, such as Bayesian program learning [8] or Long Short Term Memory neural network (LSTM, [30]). Based on these higher-level representations, *Goal Selection* forms the basis of goal-oriented behavior by selecting valuable states to be reached, where value is provided by the *Value Prediction* module. Intrinsically-motivated methods maximizing learning progress can be applied here for an efficient exploration of the environment [31]. The selected goals are reached through *Planning*, where any adaptive method of this field can be applied [32]. The resulting action plans, learned from action-state-value tuples generated by the Adaptive layer, propagate down the architecture to modulate behavior. Finally, an addressable memory system registers the activity of the Contextual layer, allowing the persistence of the agent experience over the long term for lifelong learning abilities [33]. In psychological terms, this memory system is analog to an autobiographical memory.

These high-level cognitive processes, in turn, modulate behavior at lower levels via top-down pathways shaped by behavioral feedback. The control flow is therefore distributed within the architecture, both from bottom-up and top-down interactions between layers, as well as from lateral information processing into the subsequent layers.

*B. Expressing existing Machine Learning systems within the DAC-EAI framework*

We now demonstrate the generality of the proposed DAC-EML architecture by describing how well-known Artificial Intelligence systems can be conceptually described as sub-parts of the DAC-EAI architecture (***Figure** 2*).

We start with behavior-based robotics [20], implementing a set of reactive controllers through low-level coupling between sensors to effectors. Within the proposed framework, there are described as the lower part of the architecture, spanning the Somatic and Reactive layers (***Figure** 2*B). However, those approaches are not considering the self-regulation of *internal* variables but instead of *exteroceptive* variables, such as light quantity for example.

In contrast, top-down robotic planning algorithms [34] correspond to the right column (Action) of the DAC-EAI architecture: spanning from *Planning* to *Action Selection* and *Motor Control*, where the current state of the system is typically provided by pre-processed sensory-related information along the Reactive or Adaptive layers (***Figure** 2*C). More recent Deep Reinforcement Learning methods, such as the original Deep Q-Learning algorithm (DQN, [3]) typically span over all the Adaptive layer, They use convolutional deep networks *learning abstract representation* from pixel-level *sensing* of video game frames, Q-learning for *predicting the cumulated value* of the resulting states and competition among discrete actions as an *action selection* process (***Figure** 2*D). Still, there is no real *motor control* in this system, given that most available benchmarks operate on a limited set of discrete (up-down-left-right) or continuous (forward speed, rotation speed) actions. Not shown in ***Figure** 2*, classical reinforcement learning [28] relies on the same architecture as ***Figure** 2*D, however not addressing the *representation learning* problem, since the state space is usually pre-defined in these studies (often considering a grid world).

Several extensions based on the DQN algorithm exist. For example, intrinsically-motivated deep reinforcement learning [6] extends it with a *goal selection* mechanism (***Figure** 2*E). This extension allows solving tasks with delayed and sparse reward (e.g. Montezuma Revenge) by encouraging exploratory behaviors. AlphaGo also relies on a Deep Reinforcement Learning method (hence spanning the Adaptive layer as in the last examples), coupled with a Monte-Carlo tree search algorithm which can be conceived as a *planning* process (see also [35]), as represented in ***Figure** 2*F.

Another recent work, adopting a drastically opposite approach as compared to end-to-end deep learning, addresses the problem of learning highly abstract concepts from the perspective of the human ability to perform one-shot learning. The resulting model, called Bayesian Program Learning [8], relies on a priori knowledge about the primitives to learn from and about how to compose them to build increasingly abstract categories. In this sense, it is described within the DAC-EAI framework as addressing the pattern recognition problem from the perspective of *relational learning*, where primitives are causally linked for composing increasingly abstract categories (***Figure** 2*G).

Finally, the Differentiable Neural Computer [36], the successor of the Neural Turing Machine [37], couples a neural controller (e.g. based on an LSTM) with a content-addressable memory. The whole system is fully differentiable and is consequently optimizable through gradient descent. It can solve problems requiring some levels of sequential reasoning such has path planning in a subway network or performing inferences in a family tree. In DAC-EAI terms, we describe it as an implementation of the higher part of the architecture, where causal *relations* are learned from experience and selectively stored in an *addressable memory,* which can further by accessed for reasoning or *planning* operations (***Figure** 2*H).

An interesting challenge with such an integrative approach is therefore to express a wide range of Artificial systems within a unified framework, facilitating their description and comparison in conceptual terms.

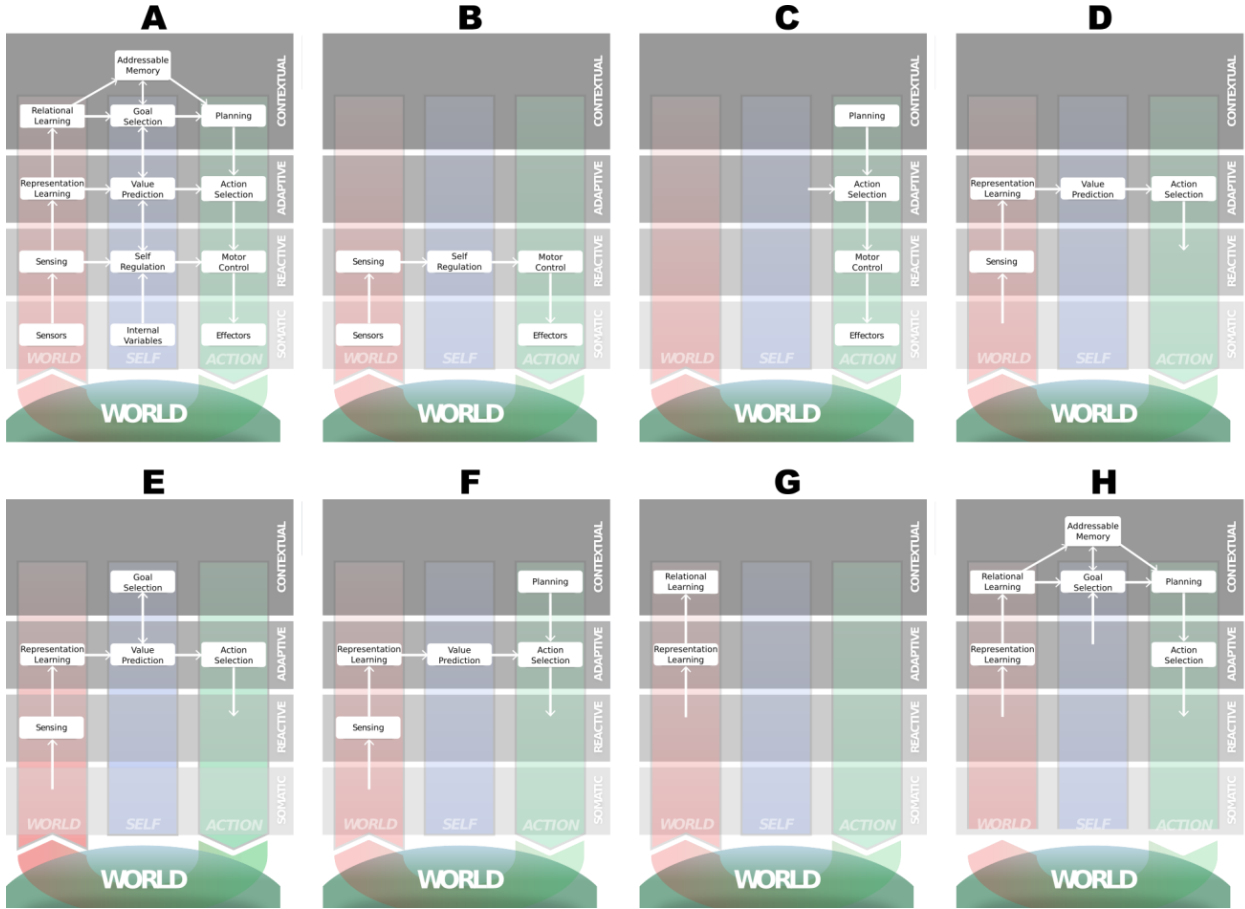

**Figure 2: The DAC-EAI architecture allows a conceptual description of many Artificial Intelligence systems within a unified framework.** A) The complete DAC-EAI architecture (see Figure 1 for a larger version). The other subfigures (B to E) show conceptual descriptions of different Artificial Intelligence systems within the DAC-EAI framework. B): Behavior-based Robotics [20]. C) Top-down robotic planning [34]. D) Deep Q-Learning [3]. E) Intrinsically-Motivated Deep Reinforcement Learning [6]. F) AlphaGo [4]. G) Bayesian Program Learning [8]. H) Differentiable Neural Computer [36].

## III. THE COGNITIVE ARMS RACE: REPRODUCING ECOLOGICALLY-VALID CONDITIONS FOR DEVELOPING COGNITIVE COMPLEXITY

A general-purpose cognitive architecture for Artificial Intelligence, as the one proposed in the previous section, tackles the challenge of general-purpose intelligence with the aim of addressing any kind of task. Traditional benchmarks, mostly based on datasets or on idealized reinforcement learning tasks, are progressively becoming obsolete in this respect. There are two reasons for this. The first one is that state-of-the-art learning algorithms are now achieving human performance in an increasing number of these traditional benchmarks (e.g. visual classification, video or board games). The second reason is that the development of complex cognitive systems is likely to depend on the complexity of the environment they evolve in[1]. For these two reasons, Machine Learning benchmarks have recently evolved toward first-person 3D game platforms embedding realistic physics [10], [11] and likely to become the new standards in the field.

It is therefore fundamental to figure out what properties of the environment act as driving forces for the development of complex cognitive abilities in embodied agents. We propose in this paper the concept of a cognitive arms race as a fundamental driving force catalyzing the development of cognitive complexity. The aim is to reproduce ecologically-valid conditions among embodied agents forcing them to continuously improve their cognitive abilities in a dynamic multi-agent environment. In natural science, the concept of an evolutionary arms race has been defined as follows: "*an adaptation in one lineage (e.g. predators) may change the selection pressure on another lineage (e.g. prey), giving rise to a counter-adaptation*" [38]. This process produces the conditions of a positive feedback loop where one lineage pushes the other to better adapt and vice versa. We propose that such a positive feedback loop is a key driving force for achieving an important step towards the development of machine general intelligence.

A first step for achieving this objective is the computational modeling of two populations of embodied cognitive agents,

---

[1] See also https://deepmind.com/blog/open-sourcing-deepmind-lab/: *"It is possible that a large fraction of animal and human intelligence is a direct consequence of the richness of our environment, and unlikely to arise without it"*.

preys and predators, each agent being driven by the cognitive architecture proposed in the previous section. Basic survival behaviors are implemented as sensorimotor control loops operating in the Reactive layer, where predators hunt preys, while preys escape predators and are attracted to other food sources. Since these agents adapt to environmental constraints through learning processes occurring in the upper levels of the architecture, they will reciprocally adapt to each other. A cognitive adaptation (in term of learning) of members of one population will perturb the equilibrium attained by the others for self-regulating their own internal variables, forcing them to re-adapt in consequence. This will provide an adequate setup for studying the conditions of entering in a cognitive arms race between populations, where both reciprocally improve their cognitive abilities against each other.

A number of previous works have tackled the challenge of solving social dilemmas in multi-agent simulations (see e.g. [13] for a recent attempt using Deep Reinforcement Learning). Within these works, the modeling of wolf-pack hunting behavior ([13], [39], [40]) is of particular interest as a starting point for bootstrapping a cognitive arms race. Such behaviors are based both on competition between the prey and the wolf group, as well as cooperation between wolves to maximize hunting efficiency. This provides a complex structure of co-dependencies among the considered agents where adaptations of one's behavior will have consequences on the equilibrium of the entire system. Such complex systems have usually been studied in the context of Evolutionary Robotics [41] where co-adaptation is driven by a simulated Darwinian selection process. However complex co-adaptation can also be studied through coupled learning among agents endowed with the cognitive architecture presented in the previous section.

It is interesting to note that there exist precursors of this concept of an arms race in the recent literature under a quite different angle. An interesting example is a Generative Adversarial Network [42], where a pattern generator and a pattern discriminator compete and adapt against each other. Another example is the AlphaGo program [4] which was partly trained by playing games against itself, consequently improving its performance in an iterative way. Both these systems owe their success in part to their ability to enter into a positive feedback loop of performance improvement.

## IV. CONCLUSION

Building upon recent advances in Artificial Intelligence and Machine Learning, we have proposed in this paper a cognitive architecture, called DAC-EAI, allowing the conceptual description of many Artificial Intelligence systems within a unified framework. Then we have proposed the concept of a cognitive arms race between embodied agent population as a potentially powerful driving force for the development of cognitive complexity.

We believe that these two research directions, summarized by the keywords *integration* and *embodiment*, are key challenges for leveraging the recent advances in the field toward the achievement of General Artificial Intelligence. This ambitious objective requires a cognitive architecture autonomously and continuously optimizing its own behavior through embodied interaction with the world. This is, however, not a sufficient condition for an agent to continuously learn increasingly complex skills. Indeed, in an environment of limited complexity with sufficient resources, the agent will rapidly converge towards an efficient strategy and there will be no need to further extend the repertoire of skills. However, if the environment contains other agents competing for the same, limited resources, the efficiency of one's strategy will depend on the strategies adopted by the others. The constraints imposed by such a multi-agent environment with limited resources are likely to be a crucial factor in bootstrapping a positive-feedback loop of continuous improvement through competition among the agents, as described in the previous section.

The main lesson of our integrative effort at the cognitive level, as summarized in Figure 2, is that powerful algorithms and control systems are existing which, taken together, span all the relevant aspects of cognition required to solve the problem of General Artificial Intelligence[2]. We see however that there is still a considerable amount of work to be done to integrate all the existing subparts into a coherent and complete cognitive system. This effort is central to the research program of our group and we have already demonstrated our ability to implement a complete version of the architecture (see [16], [24] for our most recent contributions).

As we already noted in previous publications [15], [26], [43]–[45], there is, however, a missing ingredient in these systems preventing them to being considered at the same level as animal intelligence: they are not facing the constraint of the massively multi-agent world in which biological systems evolve. We propose here that a key constraint imposed by a multi-agent world is the emergence of positive feedback loops between competing agent populations, forcing them to continuously adapt against each other.

Our approach is facing several important challenges. The first one is to leverage the recent advances in robotics and machine learning toward the achievement of general artificial intelligence, based on the principled methodology provided by the DAC framework. The second one is to provide a unified theory of cognition [46] able to bridge the gap between computational and biological science. The third one is to understand the emergence of general intelligence within its ecological substrate, i.e. the dynamical aspect of coupled physical and cognitive systems.

ACKNOWLEDGMENT

Work supported by ERC's CDAC project: "Role of Consciousness in Adaptive Behavior" (ERC-2013-ADG 341196) & EU project Socialising Sensori-Motor Contingencies (socSMC-641321—H2020-FETPROACT-2014), as well as the INSOCO Plan Nacional Project (DPI2016-80116-P).

---

[2] But this does not mean those aspects are sufficient to solve the problem.